# Uncertainty Estimation and Out-of-Distribution Detection for Deep Learning-Based Image Reconstruction using the Local Lipschitz

Danyal F. Bhutto, Bo Zhu, Jeremiah Z. Liu, Neha Koonjoo, Hongwei B. Li, Bruce R. Rosen, and Matthew S. Rosen

*Abstract*— Accurate image reconstruction is at the heart of diagnostics in medical imaging. Supervised deep learning-based approaches have been investigated for solving inverse problems including image reconstruction. However, these trained models encounter unseen data distributions that are widely shifted from training data during deployment. Therefore, it is essential to assess whether a given input falls within the training data distribution for diagnostic purposes. Uncertainty estimation approaches exist but focus on providing an uncertainty map to radiologists, rather than assessing the training distribution fit. In this work, we propose a method based on the local Lipschitz-based metric to distinguish out-of-distribution images from in-distribution with an area under the curve of 99.94%. Empirically, we demonstrate a very strong relationship between the local Lipschitz value and mean absolute error (MAE), supported by a high Spearman's rank correlation coefficient of 0.8475, which determines the uncertainty estimation threshold for optimal model performance. Through the identification of false positives, the local Lipschitz and MAE relationship was used to guide data augmentation and reduce model uncertainty. Our study was validated using the AUTOMAP architecture for sensor-to-image Magnetic Resonance Imaging (MRI) reconstruction. We compare our proposed approach with baseline methods: Monte-Carlo dropout and deep ensembles, and further analysis included MRI denoising and Computed Tomography (CT) sparse-to-full view reconstruction using UNET architectures. We show that our approach is applicable to various architectures and learned functions, especially in the realm of medical image reconstruction, where preserving the diagnostic accuracy of reconstructed images remains paramount.

*Index Terms*—Deep Learning, image reconstruction, magnetic resonance imaging, computed tomography, uncertainty estimation, local Lipschitz

## I. INTRODUCTION

FROM cancer staging and treatment planning to the functional observation of fetal cardiac health, medical imaging is an indispensable necessity for healthcare. With specialized imaging sensors, we can peer inside the human body and for instance follow a disease progression. Depending on the imaging modality, the nature of sensor data varies widely, such as ultrasound element-time space data, X-ray and Computed Tomography (CT) radon sinogram data, and Magnetic Resonance Imaging (MRI) k-space data. Based on the unique characteristics of their tomographic forward model encoding, the sensor-domain data is transformed into comprehensible images through a complex inverse problem-solving process known as image reconstruction. Hence, image reconstruction will often necessitate regularization to well-pose the inverse problem. Additionally, sensor-domain data contain millions of sample values, making accurate data processing a computationally demanding and intricate series of steps that mainly rely on hand-crafted analytical methods for image reconstruction.

In recent years, there have been notable efforts to apply Deep Learning (DL) techniques to image reconstruction [1]–[4]. In 2018, Zhu et al. introduced a novel approach that leveraged neural networks for domain-transform manifold learning, enabling the reconstruction of MRI data in a single pass by mapping sensor-domain data to image-domain in a data-driven fashion [5]. Their dual-domain manifold learning formalism, known as AUTOMAP, was capable of learning multiple forward-encoding schemes for MRI, including Radon projection and undersampled Fourier $k$-space, as well as

This work was supported by the National Science Foundation Graduate Research Fellowship under Grant No. DGE-1840990 and NSF NRT: National Science Foundation Research Traineeship Program (NRT): Understanding the Brain (UtB): Neurophotonics DGE-1633516NSF.

D. F. Bhutto is with the Department of Biomedical Engineering at Boston University, Boston, MA, 02215 and with the Athinoula A. Martinos Center for Biomedical Imaging, Department of Radiology, Massachusetts General Hospital, Charlestown, MA, 02129 (email: bhutto@bu.edu).

B. Zhu was with the Athinoula A. Martinos Center for Biomedical Imaging, Department of Radiology, Massachusetts General Hospital, Charlestown, MA, 02129 (email: bozhu5@gmail.com).

J. Z. Liu was with the Department of Biostatistics, Harvard University, Cambridge, MA, 02115. He is now with Google Research, Mountain View, CA, 94043 (email: jereliu@google.com).

N. Koonjoo, H. B. Li, and B. R. Rosen are with the Harvard Medical School, Boston, MA, 02115 and the Athinoula A. Martinos Center for Biomedical Imaging, Department of Radiology, Massachusetts General Hospital, Charlestown, MA, 02129 (email: nkoonjoo@mgh.harvard.edu, holi2@mgh.harvard.edu, and brrosen@mgh.harvard.edu).

M. S. Rosen is with Harvard Medical School, Boston, MA, 02115, the Athinoula A. Martinos Center for Biomedical Imaging, Department of Radiology, Massachusetts General Hospital, Charlestown, MA, 02129 and the Department of Physics, Harvard University, Cambridge, MA, 02138, United States (email: msrosen@mgh.harvard.edu).

reconstruct positron emission tomography (PET) images from sinogram data. Koonjoo et al. also demonstrated that domain-specific trained models could yield significantly improved results, even when the forward-encoding model remains the same [6]. Employing DL in image reconstruction also includes enhanced signal-to-noise ratio (SNR) [5] with denoising capabilities [7] and reducing image artifacts [8]. These techniques can also effectively mitigate the challenges associated with undersampled data during reconstruction [5], [9]–[11].

With constant deployment of several DL-based image reconstruction approaches into medical facilities, their robustness [12]–[15] and their applicability to data beyond the training distribution [16], [17] has been questioned. There is also a growing focus on quantifying hallucinations or false structures introduced by these techniques [18], which could potentially lead to misdiagnosis. Recent work has addressed the challenge of uncertainty estimation by training bifurcated networks that produce both a reconstructed image and a prediction of potential errors within the reconstructed image [19], [20]. Another study utilized a total deep variation regularizer in a multi-step process to reconstruct predictions and quantified pixel-wise epistemic uncertainty, type of uncertainty from the training dataset that often is reduced by more data [21]. While these methods offer a visual prediction of errors in the reconstructed image, their effectiveness is contingent on the training dataset and may not extend well to out-of-distribution (OOD) images, such as trained on brain scans while reconstructing and predicting uncertainty for OOD knee $k$-space input. In another study, researchers introduced slight inconsistencies into the training datasets by applying a constant bias to create models with minor shifts, subsequently using the variations in predictions from these models to measure epistemic uncertainty. While this approach outperformed baseline methods, the authors acknowledged its limitations in not mitigating biases introduced by the training data, which, in turn, affect the reliability of the uncertainty estimates [22]. Some traditional techniques include Monte-Carlo dropout [23] and deep ensemble [24]. However, the main limitations are the need to inference multiple times for each input for Monte-Carlo and to train multiple models for deep ensemble. Particularly in medical imaging where image sizes are large, these two approaches are computationally heavy and provide uncertainty results related to the intensity and texture of a reconstructed image, rather than uncertainty related to accuracy.

Lipschitz-based metrics to demonstrate robustness have been explored in various studies for classification to vector-to-vector regression neural network approaches [6], [25]–[33]. Weng et al. introduced a robustness metric called Cross Lipschitz Extreme Value for Network Robustness (CLEVER). They provide theoretical justification for using the local Lipschitz to estimate robustness and provide experimental results of robustness against adversarial noise [25]. Zou et al. computed the Lipschitz bounds of convolutional neural networks, including models like AlexNet and GoogleNet, illustrating how Lipschitz bounds can serve as nonlinear discriminants in classification tasks [29]. Moreover, the Lipschitz value can also be used to establish upper bounds on estimated regression error in the presence of noise, particularly in the context of vector-to-vector regression, owing to the Lipschitz continuity property of the Mean Absolute Error (MAE) [33]. It has also been employed in domain-transform manifold learning for image reconstruction, to demonstrate the robustness of AUTOMAP models [6].

Numerous neural network architectures have been proposed in the literature for DL in image reconstruction, including AUTOMAP [5], dAUTOMAP [34], RIM-net [35], XPDNet [36], and DAGAN [37]. Convolutional Neural Networks (CNNs) have also been employed for reconstruction tasks due to their capacity to reduce artifacts, preserve structural details, and denoise images [10], [38]–[41]. In our study, we shall mainly employ AUTOMAP for our analysis due to the presence of both fully connected layers and convolutional layers within the neural network architecture. The fully connected layers are designed to learn the inverse transformation, a capability supported by the universal approximation theorem [42]. Zhu et al. initially demonstrated AUTOMAP's effectiveness in reconstructing MRI $k$-space data, by learning the inverse of Fast Fourier Transform. Subsequent research has expanded its utility to various domains and functions, such as infrared interferometry for image reconstruction in astronomy [43]. We also employ UNET architectures for MRI denoising [41] and sparse-to-full view CT reconstruction [10], to highlight that our method is applicable to fully convolutional neural networks.

In this study, we showcase an approach to differentiate between out-of-distribution (OOD) and in-distribution (ID) samples that will help mitigate the risks associated with reconstructing data that falls outside the training data distribution. We present the methodology for calculating the local Lipschitz value and elaborate on the training specifics for different neural network architecture. Additionally, we introduce a reconstruction pipeline in which a radiologist has established a predefined local Lipschitz threshold for optimal performance. We provide empirical evidence supporting the use of the local Lipschitz value to estimate MAE as a means of estimating uncertainty. We compare our local Lipschitz technique to Monte-Carlo dropout and deep ensemble to demonstrate the ability to establish a performance threshold for reconstructions and to identify when an input falls outside the training data distribution, enabling the use of alternative methods for reconstruction. This approach helps avoid reconstructions in situations where the model's generalization may be limited.

## II. METHODOLOGY

### A. DL for Image Reconstruction Preliminaries

In DL-based image reconstruction tasks, the objective is to reconstruct a target image $y$ from input data $x$ by learning a reconstruction mapping $f: \mathbb{R}^{n^2} \to \mathbb{R}^{n^2}$ that minimizes the cost function $C(\hat{y}, y)$ where $\hat{y} = \hat{f}(x)$ is ground truth image and

$y = f(x)$ is the predicted image. The data is denoted as $\{y_i, x_i\}_i^n$ where $i^{th}$ observation, $x_i$ is an $n \times n$ set of input parameters, and $y_i$ is an $n \times n$ set of image output values. The accuracy can be defined between the truth vector $\hat{y}_i = \{\hat{y}_1, ..., \hat{y}_{n^2}\}$ and the predicted vector $y_i = \{y_1, ..., y_{n^2}\}$ using the Mean Absolute Error (MAE) function, $MAE(\hat{y}, y) = \frac{1}{n^2}\sum_{i=1}^{n^2} \|\hat{y}_i - y_i\|_1$.

### B. Calculation of the Local Lipschitz

By definition, a function $f$ is $L$-Lipschitz continuous if for any input $x, y \in \mathbb{R}^n$, there exists a nonnegative constant $L \geq 0$.

$$\|f(x) - f(y)\| \leq L\|x - y\| \quad (1)$$

The Lipschitz constant $L$ is the maximum ratio of variations in the output space over variations in the input space. This allows the Lipschitz constant to be a measure of the sensitivity of a Lipschitz continuous function with respect to the input perturbations. The local Lipschitz constant is calculated when $y = x + e$, where $e$ is some input perturbation such as Gaussian noise.

For our reconstruction task, let us denote $\Phi$ for the DL operation, either AUTOMAP or UNET, $L_\Phi$ for the Lipschitz constant, $x$ for input image, and $e$ for error. We rewrite (1) as follows.

$$\|\Phi(x+e) - \Phi(x)\| \leq L_\Phi \|x + e - x\| \quad (2)$$

$$\|\Phi(x+e) - \Phi(x)\| \leq L_\Phi * \|e\| \quad (3)$$

Given the small magnitude of $\|e\|$, we reorganize (3) and place an upper bound by empirically calculating $L_\Phi$, where the image $x' = x + e$.

$$L_\Phi \geq \frac{\|\Phi(x') - \Phi(x)\|}{\|x' - x\|} \quad (4)$$

$$L_\Phi \geq \frac{1}{n^2}\sum_{i=1}^{n^2} \frac{\|\Phi(x') - \Phi(x)\|}{\|x' - x\|} \quad (5)$$

Equation (4) lets us estimate the $L_\Phi$ for any image after we perturb it by Gaussian noise $e$ and compare the variations in the output space over the input space. By calculating the mean of the difference in output to the difference in input as demonstrated in (5), the $L_\Phi$ value can be interpreted as representing the impact the network has on the accuracy in recreating the output, given that the input difference remains small. For the AUTOMAP, Monte-Carlo dropout, and deep ensemble models with noise levels ranging from 5% to 20% of the standard deviation of input $k$-space intensities added as Gaussian noise. Table 1 presents Spearman's rank correlation coefficient values, highlighting a very strong correlation between the $L_\Phi$ and MAE. Since DL image reconstructions models learn denoising and lead to SNR gain, in cases where the network struggles to effectively manage noise and produces an image that significantly deviates from a clean reconstruction, the local Lipschitz constant is associated with increased MAE and uncertainty.

### C. Comparison to Baseline

Given the inherent uncertainties associated with real-world data distributions, it is imperative to identify OOD test cases, particularly where a model has not generalized effectively. For the model AUTOMAP as illustrated in Fig. 1a, we present two approaches, using the local Lipschitz and measuring the variance in output after perturbing the input with Gaussian noise four times, as demonstrated in (6) where $\Phi$ is AUTOMAP, $x$ is the input image, and $e$ is 5% Gaussian noise added.

$$Var[\Phi(x + e_1), \Phi(x + e_2), \Phi(x + e_3), \Phi(x + e_4)] \quad (6)$$

We compare these methods to baseline approaches, specifically Monte-Carlo dropout and deep ensemble. For the Monte-Carlo method, we modified the AUTOMAP architecture with the addition of a dropout layer, as illustrated in Fig. 1b. During inference, the dropout layer activates and deactivates different nodes, resulting in multiple outputs for a single input. Hence, with 50 iterations, 50 clean and 50 noisy images were reconstructed for each clean and noisy input, respectively, and the local Lipschitz was calculated using the average outputs and variance of all the 50 clean reconstructed images. For the deep ensemble method, five AUTOMAP models were trained and inferenced (see Fig. 1a) and the local Lipschitz of the average output was computed, along with the variance of the five images generated by each trained model. In Fig. 2, the receiver operating characteristic curves (ROC) and AUC values are plotted. In Fig. 3, the uncertainty estimation maps for AUTOMAP using the local Lipschitz, Monte-Carlo dropout, and deep ensemble techniques are compared.

### D. Neural Network Architectures

The AUTOMAP architecture is diagramed in Fig 1a. The first fully connected layer, FC1, has 25,000 nodes with the tanh activation function. The second fully connected layer, FC2, has 16384 nodes and no activation function. Both convolutional layers, Conv1 and Conv2, have 128 filters of size 5 x 5, with the tanh and relu activation functions respectively. The convolutional transpose layer has 1 filter of size 7 x 7 and no activation function. The AUTOMAP GitHub code, https://github.com/MattRosenLab/AUTOMAP, was used as the initial starting point and further modified. The baseline methods for comparison were Monte-Carlo and deep ensemble. For the Monte-Carlo method, a dropout layer was introduced after the Conv1 layer, with a 25% probability of randomly deactivating nodes, as illustrated in Fig. 1b. This modification yielded results that were comparable in accuracy to the single AUTOMAP model. Given that AUTOMAP is a computationally and memory-intensive architecture, five individual AUTOMAP models were trained to generate five images for the deep ensemble method.

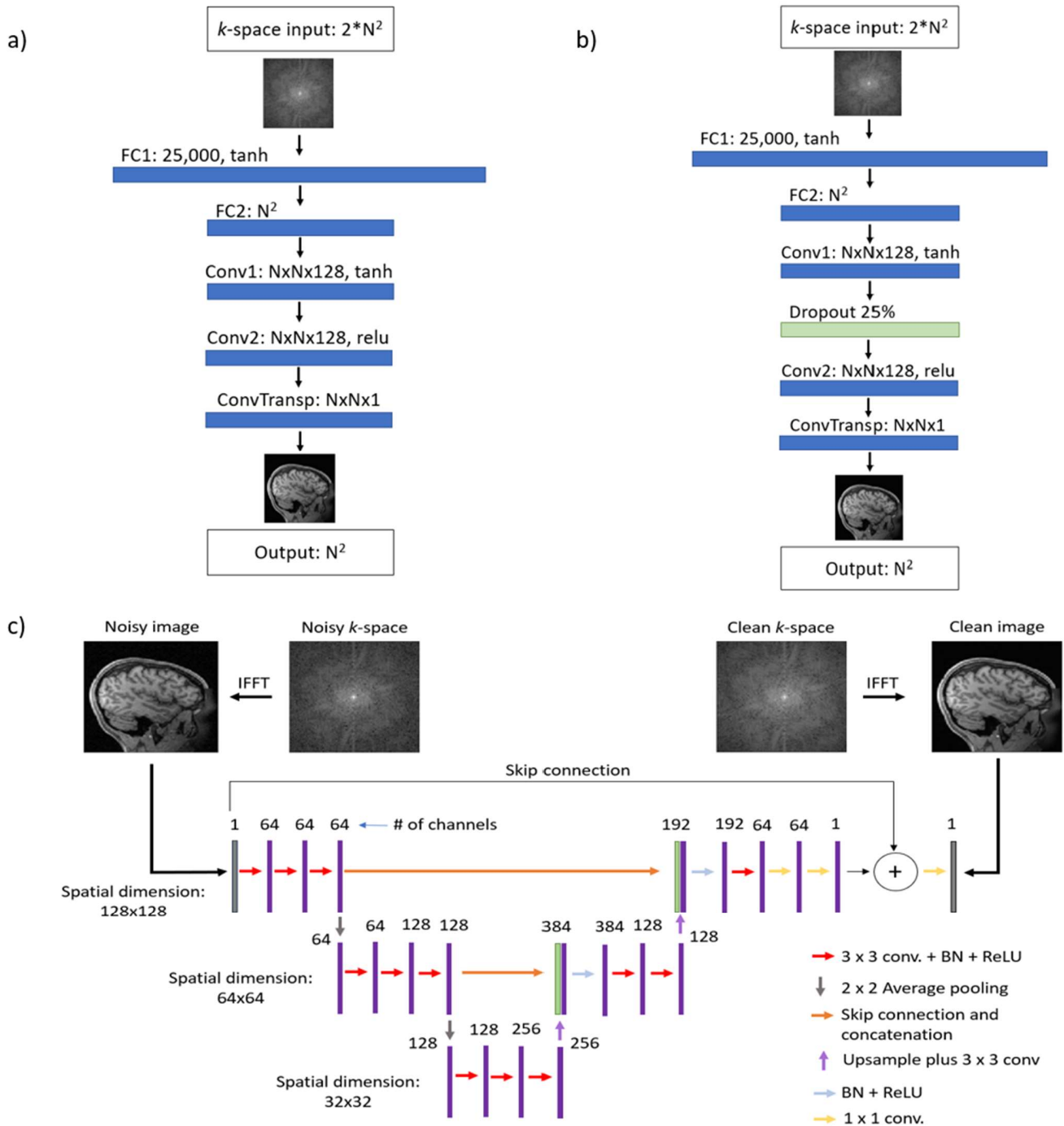

Fig. 1: a) AUTOMAP architecture utilized for the single model and deep ensemble method. b) AUTOMAP architecture with a dropout layer incorporated after the first convolutional layer for the Monte-Carlo method. c) UNET architecture for denoising MRI images. First, IFFT is performed on the noisy k-space to form the noisy image input. Then the UNET employs residual learning through skip connections to learn the difference between the noisy image input and clean image output. The same architecture is employed to transform a sparse view CT image, reconstructed from a subsampled sinogram, to a full view CT image.

Mehta *et al.* demonstrated the applicability of UNETs for denoising MRI images, which leads to improved SNR and enhanced prediction accuracy for brain tumor detection [41]. Similarly, Jin *et al.* showed that a UNET architecture can effectively learn the transformation from a sparsely sampled reconstructed CT image to a fully sampled reconstructed image [10]. Our UNET architecture was inspired by Jin *et al.*, and it incorporates skip connections to facilitate residual learning. This design enables the network to learn the difference between the input image and the ground truth. The UNET architecture is diagramed in Fig. 1c. We employed the same UNET architecture for training models to denoise MRI images and reconstruct full view from sparse view CT images.

*E. Image datasets and pre-processing*

For AUTOMAP, the training dataset consisted of 50,000 2D T1 weighted brain magnitude-only MRI images acquired at 3T from the MGH-USC Human Connectome Project (HCP) public dataset [44]. First, each image was cropped to 256 x 256 and subsampled to 128 x 128. Then each image was tiled to create a 256 x 256 image containing four reflections of the 128 x 128. Lastly, the final image was created by randomly cropping a 128 x 128 section. This approach to data augmentation was employed to promote translational invariance during training [5], [6]. The *k*-space input was generated by forward-encoding each image using the Fast Fourier Transform provided by MATLAB's native 2D FFT function. To learn a robust internal representation and promote manifold learning, the *k*-space input was noise corrupted with one percent multiplicative noise. The target image remained 'noise-free'. To prepare the data for the network, the real and imaginary values of the *k*-space input were separated and then concatenated, resulting in an input vector of size 1 x 32,768 for each image. The target image was vectorized to size 1 x 16384.

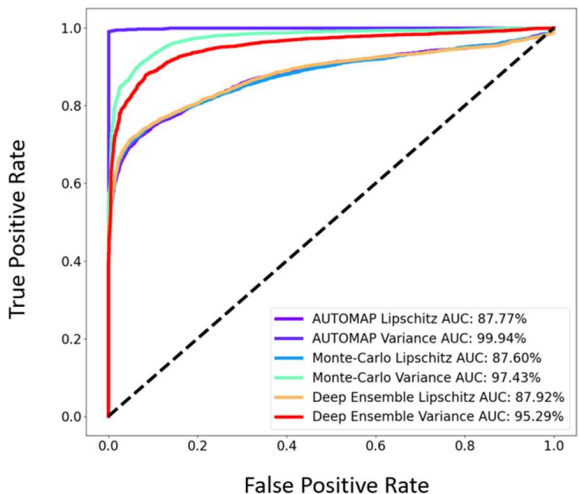

Fig. 2: ROC curves and AUC values of six methods to detect OOD knee images from ID brain images using the $L_\Phi$ values and the variance of multiple outputs. When using the $L_\Phi$ values, the single model AUTOMAP, with an AUC of 87.77%, performed comparably to the Monte-Carlo and deep ensemble methods with an AUC of 87.60% and 87.92% respectively. Using variance, the AUTOMAP outperforms both baseline methods with an AUC of 99.94% versus an AUC of 97.43% and 95.29% for Monte-Carlo and deep ensemble methods respectively.

We curated two testing datasets for analysis - an ID and an OOD datasets. In medical imaging, OOD is defined as inputs that are unseen due to a selection bias in the training data [45], [46]. The ID dataset consisted of 5,000 brain MRIs from the HCP dataset but was not a subset of the training set. The OOD dataset consisted of 5,000 knee MRI images acquired at 1.5T and 3T from the NYU fastMRI dataset [47]. While the training and ID datasets comprised of T1 weighted images, the OOD dataset comprised of coronal proton density-weighted with and without fat suppression, axial proton density-weighted with fat suppression, sagittal proton density, and sagittal T2-weighted with fat suppression images. The fastMRI datasets constitutes as OOD due to the images acquired using different field strength, MR techniques, and parameters. All fastMRI images were preprocessed and normalized using the pyFFTW, a python wrapper around the FFTW library. The ID and OOD datasets were not augmented.

Both the denoising MRI UNET and the CT sparse-to-full view reconstruction UNET utilized the 50,000 2D T1 weighted magnitude-only brain images from the MGH-USC HCP dataset for training, with the same data augmentations previously described. To form the noisy image input for the denoising MRI UNET, the *k*-space was generated using MATLAB's native 2D FFT function, 10% Gaussian noise was added to the real and imaginary values of *k*-space, and the final noisy image was reconstructed using MATLAB's native 2D IFFT function. The clean output was kept as 'noise-free' by reconstructing the image from the clean *k*-space. The test dataset for the denoising MRI UNET consisted of 5,000 brain images from the ID dataset and was preprocessed without data augmentations. To form the sparse view reconstructed CT image, each image was forward encoded using MATLAB's native Radon function resulting in a 360 view sinogram, 10% Gaussian noise was added, the sinogram was subsampled by a factor of 4 to acquire 90 views, and lastly, the sparse view image input was reconstructed using MATLAB's native Iradon function. The full view image output was reconstructed from the full 360 view noise-free sinogram using the Iradon function. The test dataset for the CT reconstruction UNET consisted of 5,000 brain images from the ID dataset and was preprocessed without data augmentations.

*F. Training details*

For training AUTOMAP, we employed the RMSProp optimization algorithm (see http://www.cs.toronto.edu/~tijmen/csc321/slides/lecture_slides_lec6.pdf) with the following hyperparameters: learning rate = .00002, momentum = 0.0, decay = 0.9 with minibatches of size = 50. We introduced weight regularization to all the fully connected and convolutional layers with an $l_2$-norm penalty ($\lambda = .001$) to prevent overfitting, not performed in the original paper [5]. An additional $l_1$-norm penalty ($\gamma = .0001$) was applied to the feature map activations of the second convolutional layer to promote sparse representations. The cost function minimized was the sum squared loss difference between the network output and the target image. All models were trained for 100 epochs. AUTOMAP typically took 7-8 hours to train using the TensorFlow 2.2.0 machine learning framework [48] with 2 NVIDIA Tesla GV100 graphics processing units (GPUs) with 32 GB memory capacity each.

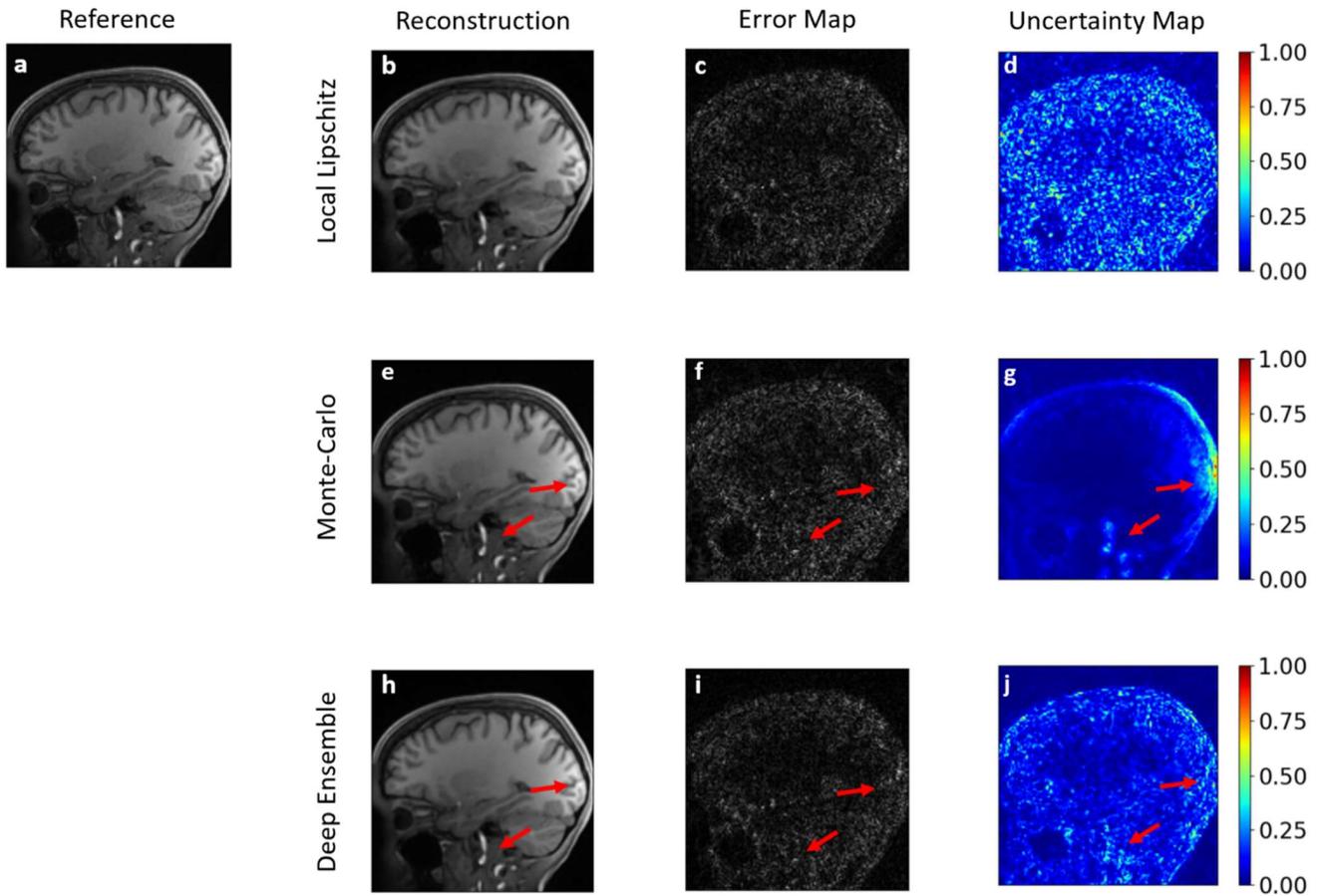

Fig. 3: MRI reconstructions with a) reference image, b-d) AUTOMAP with local Lipschitz, e-g) Monte-Carlo, and h-j) deep ensemble reconstruction, error map, and uncertainty map results. For Monte-Carlo dropout and deep ensemble respective uncertainty maps g) and j), significant variations are concentrated in areas with high intensity and noticeably pronounced texture patterns in the reconstructions e) and h) respectively, indicated by red arrows. These regions aren't correlated to high error regions as indicated by red arrows in their respective error maps f) and i). In contrast, b-d), which represents the local Lipschitz approach, calculated using 5% Gaussian noise, offers a more effective method for uncertainty estimation. This is because the noise in k-space is not localized to high or low frequencies but rather distributed throughout and modeled using Gaussian noise.

For both UNETs, we employed the RMSProp optimization algorithm with the following hyperparameters: learning rate = .0002, momentum = 0.0, decay = 0.9 with minibatches of size = 100, and trained for 100 epochs. During training, one percent multiplicative noise was also applied to the image input. The cost function minimized was sum squared loss difference between the network output and the target image. The UNETs typically took 11-12 hours to train using TensorFlow 2.2.0 machine learning framework [48] with 2 NVIDIA Tesla GV100 graphics processing units (GPUs) with 32 GB memory capacity each.

## III. RESULTS

### A. Out-of-distribution Detection

To guarantee only the test data that falls within the training distribution is used for diagnosis, we validate our suggested Lipschitz and variance metrics for OOD detection against traditional approaches. As depicted in Fig. 2, the plots of the ROC curves and AUC values for six methods aimed at detecting OOD knee images from the ID brain images using the $L_\Phi$ values and the variance of multiple outputs. We compared the single AUTOMAP model with the baseline methods of Monte-Carlo and deep ensemble. When using the $L_\Phi$ values, AUTOMAP, with an AUC of 87.77%, performs similarly to the Monte-Carlo and deep ensemble methods, which achieve

TABLE 1
SPEARMAN'S RANK CORRELATION COEFFICIENT VALUES FOR THE RELATIONSHIP BETWEEN $L_\Phi$ AND MAE

| Gaussian Noise Level | Spearman's rank correlation coefficients between $L_\Phi$ and MAE | | |
|---|---|---|---|
| | Single Model AUTOMAP | Monte-Carlo | Deep Ensemble |
| 5% | 0.8475 | 0.8687 | 0.8578 |
| 10% | 0.8364 | 0.8566 | 0.8492 |
| 15% | 0.8222 | 0.8420 | 0.8363 |
| 20% | 0.8088 | 0.8248 | 0.8171 |

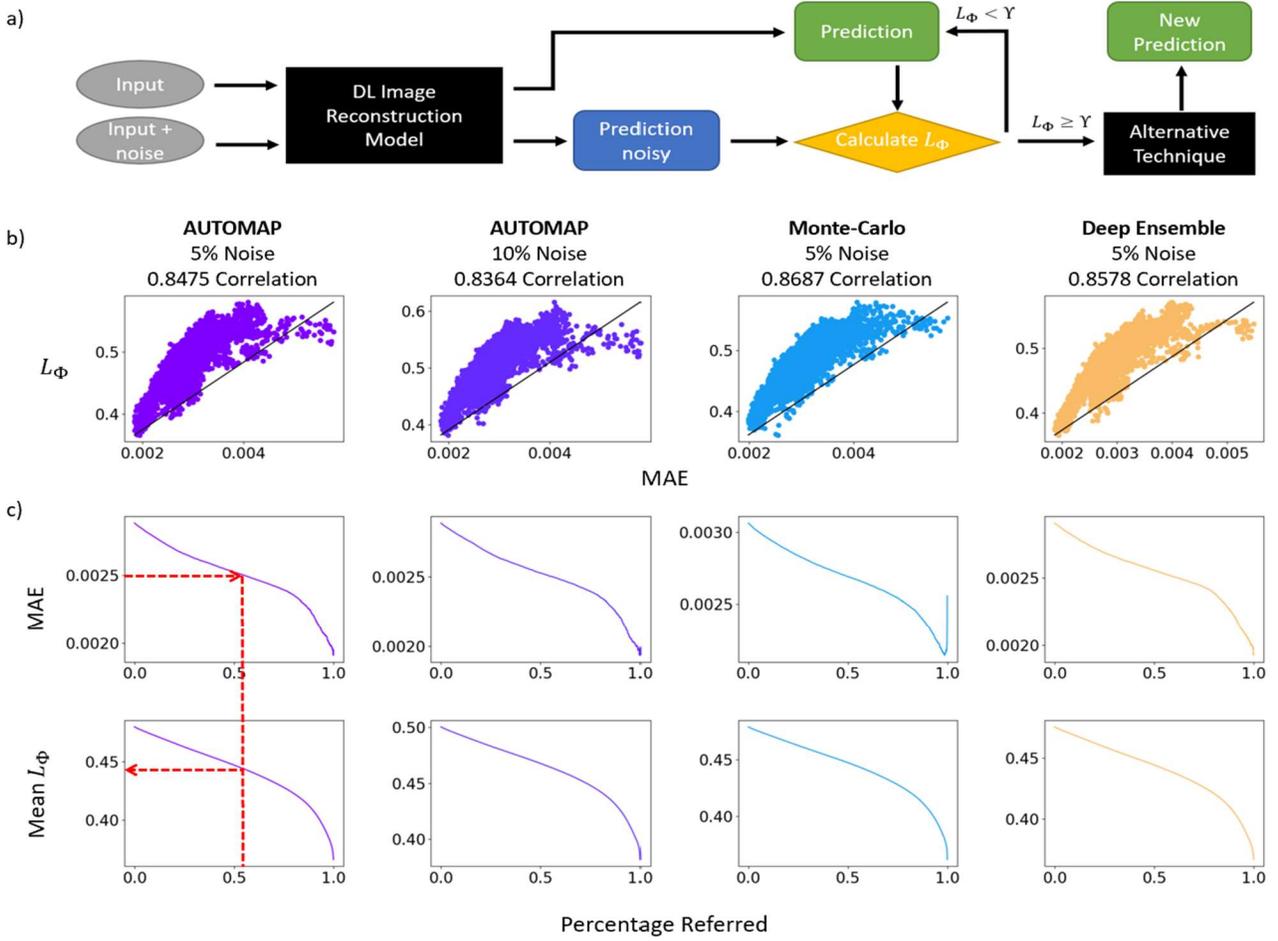

Fig. 4: a) Image reconstruction pipeline designed for integration into clinical settings to assess suitable reconstruction performance through calculating the local Lipschitz value, $L_\Phi$. If the $L_\Phi$ is below a predetermined threshold $\Upsilon$, the DL image reconstruction model performed with sufficient accuracy. If the $L_\Phi$ is above the threshold $\Upsilon$, there is greater uncertainty than desired, and the data should be processed using an alternative technique. b) Empirical evidence of the monotonic relationship between the local Lipschitz value, $L_\Phi$, and MAE for 5,000 ID images with 5% and 10% Gaussian noise for the single AUTOMAP model, as well as for Monte-Carlo and deep ensemble methods, each with 5% noise, left to right columns respectively. The Spearman's rank correlation coefficient values are listed on the top and suggest a very strong correlation. c) Selective prediction to determine threshold $\Upsilon$. As indicated by the red arrows in the leftmost column, a radiologist can determine the upper threshold for MAE that qualifies as suitable performance and subsequently, by matching the percentage of referred images, identify the threshold $\Upsilon$ from the mean $L_\Phi$ plot.

AUCs of 87.60% and 87.92%, respectively. When employing variance, AUTOMAP outperforms both baseline methods with an impressive AUC of 99.94%, compared to AUCs of 97.43% and 95.29% for Monte-Carlo and deep ensemble methods, respectively.

### B. The Local Lipschitz as Uncertainty Estimation Metric

The applicability of the local Lipschitz value was extended to its utility in estimating uncertainty. We demonstrate the empirical evidence of its correlation with MAE, and establish an uncertainty threshold for accuracy. Table 1 displays the Spearman's rank correlation coefficient values for three reconstruction methods, each calculated using the ID dataset consisting of 5,000 images. These methods include the single model AUTOMAP, Monte-Carlo dropout, and deep ensemble. Table 1 shows a remarkably strong correlation between the $L_\Phi$ and MAE for all three methods across added Gaussian noise levels ranging from 5% to 20%, highlighting the effectiveness of using the $L_\Phi$ as an indicator of reconstruction quality and uncertainty.

In Fig. 3, we illustrate MRI reconstructions with a) reference image, b-d) AUTOMAP with local Lipschitz, e-g) Monte-Carlo, and h-j) deep ensemble reconstruction, error map, and uncertainty map results. In the case of Monte-Carlo and deep ensemble respective uncertainty maps g) and j), significant variations are concentrated in areas with high intensity and noticeably pronounced texture patterns in the reconstructions e) and h) respectively, indicated by red arrows. These regions aren't correlated to high error regions as indicated by red arrows in their respective error maps f) and i). In contrast, b-d), which represents the local Lipschitz approach, calculated using 5% Gaussian noise, offers a more effective method for uncertainty estimation. This is because the noise in $k$-space is not localized to high or low frequencies but rather distributed throughout and modeled using Gaussian noise.

In Fig. 4a, we present a pipeline designed for integration into clinical settings to assess suitable reconstruction performance. The algorithm deployed in this pipeline reconstructs two images, one noisy and one clean, to calculate the local Lipschitz value, $L_\Phi$. This value is obtained by measuring the difference between the output and input and then compared to a predetermined uncertainty threshold, $\Upsilon$. If $L_\Phi < \Upsilon$, it indicates that the DL image reconstruction model performed with sufficient accuracy. However, if $L_\Phi \geq \Upsilon$, the data should be processed using an alternative technique. This approach helps ensure that only images reconstructed with adequate accuracy are retained for further analysis and diagnosis purposes, which is of the utmost importance in clinical settings.

In Fig. 4b, we provide a visual representation of the relationship between the $L_\Phi$ for 5,000 ID images, using the single model AUTOMAP with 5% and 10% noise, as well as Monte-Carlo and deep ensemble methods, each with 5% noise. This correlation reinforces the utility of the local Lipschitz as a measure of reconstruction quality and uncertainty.

In Fig. 4c, we illustrate how a threshold, $\Upsilon$, can be determined through selective prediction for reconstruction methods. We arrange the $L_\Phi$ values of the ID validation dataset in descending order. Next, we progressively 'refer' images with the highest $L_\Phi$ to an alternative reconstruction technique, and at each step, calculate the mean $L_\Phi$ and the MAE of the remaining set. As the top and bottom rows of Fig. 4c show, as more images are referred during each iteration, both the mean $L_\Phi$ and the MAE of the remaining images decrease. As indicated by the red arrows in the leftmost column, a radiologist can determine the upper threshold for MAE that qualifies as suitable performance and subsequently, by matching the percentage of referred images, identify the threshold $\Upsilon$ from the mean $L_\Phi$ plot.

To demonstrate that this monotonic relationship extends to other architectures and learned functions, in Fig. 5a, we observe that the same relationship holds for an MRI denoising UNET, composed entirely of convolutional layers. There is a very strong correlation, indicated by the Spearman correlation value of 0.7113, between the $L_\Phi$ and MAE. This network was trained to denoise reconstructions with 10% noise, and we calculated the $L_\Phi$ by taking the difference between a reconstruction with 15% noise and one with 10% noise. In Fig. 5b, we present the $L_\Phi$ versus MAE plot for the sparse-to-full view CT UNET reconstruction model. Once again, we calculated the $L_\Phi$ by taking the difference between a reconstruction with 15% noise and one with 10% noise. The same monotonic relationship is evident, with a very strong Spearman correlation value of 0.7722.

This evidence strongly suggests that the relationship between the $L_\Phi$ and MAE can serve as an effective method for estimating uncertainty and can be applied as a checkpoint in the reconstruction pipeline for various architectures and learned functions. This approach ensures that only images reconstructed with sufficient accuracy are used in diagnostic applications, enhancing the reliability and trustworthiness of the models deployed.

### C. Identification of False Positives (FP) to Guide Data Augmentation and Reduce Epistemic Uncertainty

In Fig. 6a, we display the $L_\Phi$ vs MAE plot for 5,000 OOD knee images. These $L_\Phi$ values were calculated by applying 5% Gaussian noise. Due to the moderate relationship indicated by the Spearman's correlation value of 0.4362 between the $L_\Phi$ and MAE, we investigated the specific samples that exhibited low $L_\Phi$ and high MAE. The red box highlights instances classified as FP with low $L_\Phi$ ($L_\Phi < .6$) and high MAE ($MAE > .023$). In Fig. 6b, from the top row to the bottom row, we display the FP images, random images from the OOD dataset, and random images from the ID dataset, representing the training dataset, respectively. The FP images are of high intensity and encompass the entire frame with a different field-of-view (FOV) compared to the random OOD or random ID images. This information unveils a weakness in our existing data augmentation pipeline and provides valuable information for reducing epistemic uncertainty leading to improved performance and generalization.

### IV. DISCUSSION

In this study, we emphasize the importance of identifying if an image is ID and reconstructed with minimal uncertainty, ensuring that patient safety is never compromised. We present a single model AUTOMAP variance method for detecting OOD images, which surpasses baseline methods in performance while remaining simple to implement. It's important to note that the baseline methods require a significantly larger amount of memory and time. The Monte-Carlo method demands multiple inferences for each input, involving 50 inferences and storage of 50 images per input. The deep ensemble method necessitates the training and storage of multiple models, specifically five AUTOMAP models, which results in five inferences and storage of five images per input. The limitation in our OOD detection method is the need to inference with noise multiple times for variance calculation (n=4), which is still considerably less computationally intensive.

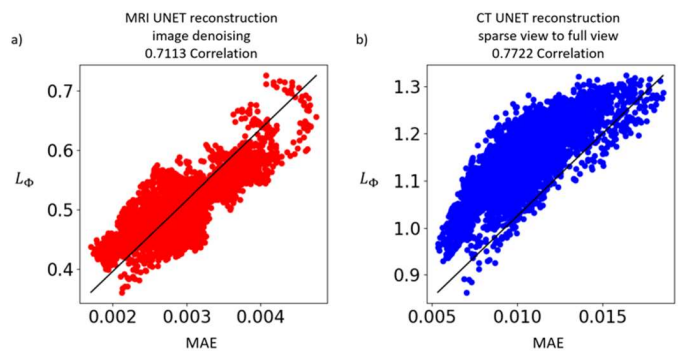

Fig. 5: a) The local Lipschitz value, $L_\Phi$, versus MAE plot for the denoising MRI UNET indicates a strong correlation with a Spearman's rank correlation coefficient value of 0.7113. b) The local Lipschitz value, $L_\Phi$, versus MAE plot for the CT sparse-to-full view reconstruction UNET indicates a strong correlation with a Spearman's rank correlation coefficient value of 0.7722. These plots empirically demonstrate the monotonic relationship between the $L_\Phi$ and MAE and that the local Lipschitz uncertainty estimation method can apply to different architectures and learned functions.

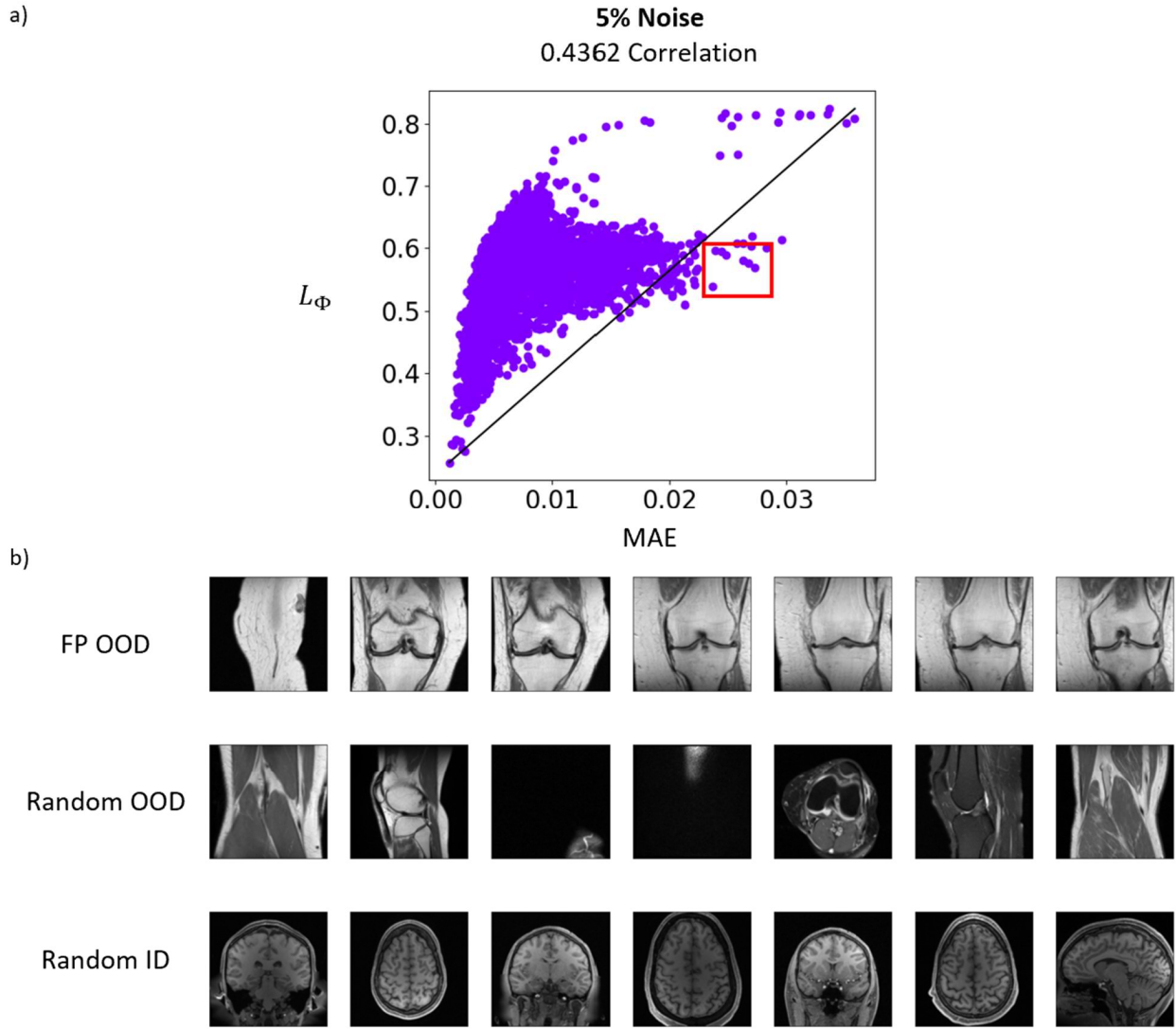

Fig. 6: The local Lipschitz value, $L_\Phi$, versus MAE plot for OOD knee images using the AUTOMAP reconstruction model. The red box indicates the FPs with low $L_\Phi$ ($L_\Phi < .6$) and high MAE ($MAE > .023$). b) from the top row to the bottom row, FP images, random images from the OOD dataset, and random images from the ID dataset that represent the training dataset, respectively. The FP images are of high intensity and encompass the entire frame with a different FOV than the random OOD or random ID images. This information unveils a weakness in our existing data augmentation pipeline and provides valuable information for reducing epistemic uncertainty leading to improved performance and generalization.

Furthermore, we demonstrate a monotonic relationship between the local Lipschitz value and MAE, as measured by the Spearman's rank correlation coefficient, and expand its use from studying robustness to uncertainty estimation of DL image reconstruction techniques. We show results for both AUTOMAP and UNET architectures and for different reconstruction applications and methods. Through selective prediction, we illustrate a method for determining a local Lipschitz uncertainty threshold. When the local Lipschitz value of an image exceeds this threshold, the reconstructed image is deemed inaccurate, and an alternative technique should be employed for reconstruction. In the future, we aspire to collaborate with radiologists to determine the optimal local Lipschitz threshold for a deployed model in a clinical setting. Lastly, we use our local Lipschitz method to identify FP and provide potential improvements to the data augmentation pipeline that may help model generalization. Given the expanding role of AI in medical image reconstruction, with companies like GE Healthcare gaining FDA clearance for their models, it has become increasingly crucial to establish accuracy checkpoints within the reconstruction pipeline. These checkpoints will help ensure patient safety while maintaining high standards of accuracy and reliability.